\documentclass[11pt]{article}

\usepackage[preprint]{acl}

\usepackage{times}
\usepackage{latexsym}
\usepackage{amsmath}
\usepackage{algorithm}  
\usepackage{algorithmic}  
\usepackage{multirow}
\usepackage{arydshln}
\usepackage{booktabs}
\usepackage{fancyvrb}
\usepackage[T1]{fontenc}
\usepackage{CJKutf8}

\usepackage[utf8]{inputenc}

\usepackage{microtype}

\usepackage{inconsolata}

\usepackage{graphicx}

\usepackage{caption}
\title{\textbf{S}elf-\textbf{S}peculative \textbf{B}iased \textbf{D}ecoding for Faster Re-translation}

\author{
 \textbf{Linxiao Zeng\textsuperscript{1}},
 \textbf{Haoyun Deng\textsuperscript{1}},
 \textbf{Kangyuan Shu\textsuperscript{1}},
 \textbf{Shizhen Wang\textsuperscript{1}}
\\
 \textsuperscript{1}Zoom Communications, Inc
\\
  \texttt{firstname.lastname@zoom.com} \\
}

\begin{document}
\maketitle
\begin{abstract}
    Large language models achieve strong machine translation quality but incur high inference cost and latency, posing challenges for simultaneous translation. Re-translation provides a practical solution for off-the-shelf LLMs by repeatedly regenerating the target output as the source input grows, but it suffers from substantial redundant computation. We propose Self-Speculative Biased Decoding (SSBD), a simple and tuning-free inference method that accelerates re-translation by exploiting temporal coherence in streaming translation. SSBD reuses the model’s previous output as a speculative draft for the updated input, verifies the draft efficiently in a single forward pass with a lightweight bias, and resumes autoregressive decoding only from the first divergence. We further introduce a display-only masking strategy that hides unstable suffixes from the user interface while retaining them in the draft for verification and potential acceptance. Experiments show that SSBD achieves substantial speedup over standard re-translation while maintaining comparable translation quality, without architectural changes, auxiliary models, or extra fine-tuning.
\end{abstract}

\begin{figure}[t]
  \includegraphics[width=\columnwidth]{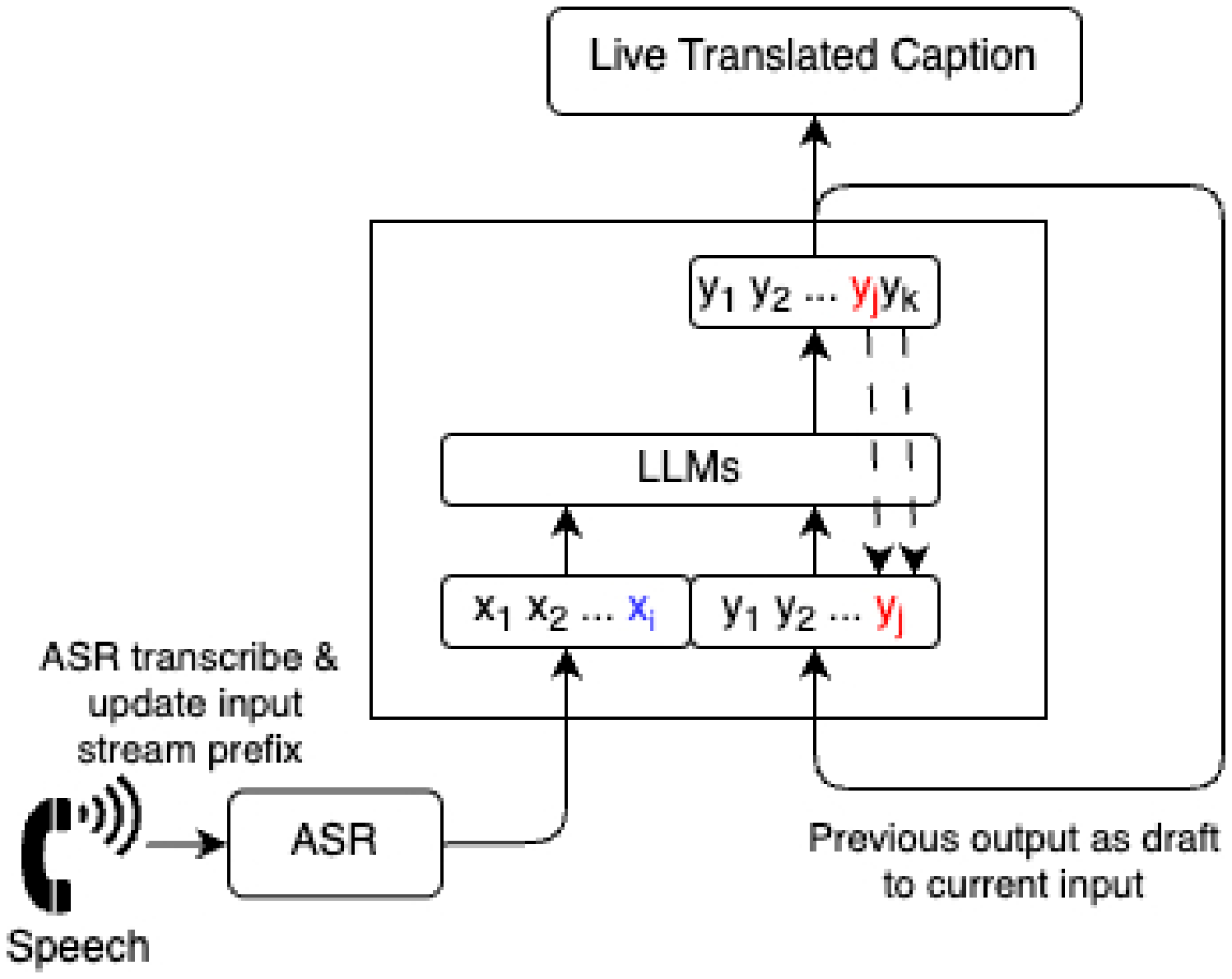}
  \caption{Self-speculative decoding for live captioning: reuse the previous translation as draft, verify in parallel, then resume from the first divergence.}
  \label{fig:draft}
\end{figure}

\section{Introduction}
LLMs have recently demonstrated strong performance on machine translation tasks \citep{xu2024a}, making them an appealing option for real-world deployment \citep{lyu-etal-2024-paradigm, pang-etal-2025-salute}. However, their large model size increases inference cost and latency, posing a major challenge for online and streaming applications such as simultaneous translation.

Simultaneous machine translation (SimulMT) requires generating partial translations before the full source sentence is available, balancing translation quality and latency. While policy-based approaches such as wait-$k$ \citep{ma-etal-2019-stacl} offer fine-grained control over read–write decisions, they typically rely on specialized architectures or additional training.
This limits their applicability to off-the-shelf LLMs. As a result, re-translation \citep{arivazhagan-etal-2020-translation} has emerged as a practical solution for simultaneous translation. Re-translation repeatedly regenerates the target prefix as the source input grows, achieving strong translation quality but incurring substantial redundant computation. This inefficiency becomes particularly severe when using large autoregressive models.


In this work, we propose \textit{Self-Speculative Biased Decoding} (SSBD), a simple and tuning-free decoding strategy that accelerates re-translation for off-the-shelf LLMs.
Unlike existing speculative decoding methods, SSBD is explicitly designed for streaming scenarios where successive inputs are strongly correlated.
SSBD exploits the temporal coherence inherent in streaming translation: the translation produced at the previous streaming step is reused as a speculative draft for the current input and verified efficiently in parallel. 
To further improve draft acceptance and output stability, we introduce a biased verification mechanism that softly favors draft tokens during verification, increasing prefix reuse while still allowing corrections as more source context becomes available. We additionally apply a display-only mask-$k$ heuristic to reduce perceptible flickering without sacrificing decoding efficiency.

Experiments on simultaneous translation benchmarks demonstrate that SSBD achieves up to 1.7$\times$ speedup over standard autoregressive re-translation while maintaining comparable translation quality and reducing flickering. Our method requires no architectural changes, auxiliary models, or fine-tuning, making it readily applicable to existing LLM-based streaming translation systems.

\section{Related Work}
\paragraph{Simultaneous Machine Translation.}
Simultaneous machine translation (SimulMT) aims to generate translations in an online manner while balancing translation quality and latency. Existing work largely follows two paradigms: policy-based methods and re-translation. Policy-based approaches explicitly model read-write decisions, using fixed strategies such as wait-$k$ \citep{ma-etal-2019-stacl} and learned adaptive policies \citep{gu-etal-2017-learning} or monotonic attention mechanisms \citep{arivazhagan-etal-2019-monotonic}, but often require specialized architectures or fine-tuning, limiting applicability to off-the-shelf LLMs.
In contrast, re-translation-based methods repeatedly re-decode the target prefix as new source tokens arrive and have been shown to perform comparably to, or even better than, dedicated streaming systems \citep{arivazhagan-etal-2020-translation, 9054585}. This repeated decoding, however, introduces substantial redundant computation, resulting in high latency and computational cost as model size grows, particularly for LLMs.

\paragraph{Speculative Decoding.}
Speculative decoding accelerates autoregressive generation by verifying multi-token drafts in parallel \citep{leviathan2023fastinferencetransformersspeculative, chen2023acceleratinglargelanguagemodel}.
Existing methods typically rely on auxiliary draft models \citep{xia-etal-2023-speculative, xia2025swift} or architectural modifications within the target model \citep{zhang-etal-2024-draft, cai2024medusasimplellminference}.
These approaches are primarily designed for offline decoding and often require training or model changes, making them less suitable for streaming.

\begin{table*}[ht]
\centering
\renewcommand{\arraystretch}{0.85}
\begin{tabular}{l|l|l}
 \hline
 \textbf{No.} & \textbf{Input} & \textbf{Output} \\
 \hline
 1 & This is & C'est \\
 2 & This is an example & C'est un example \\
 3 & This is an example of self-speculative & C'est un example d'auto-spéculation \\
 4 & This is an example of self-speculative decoding. & C'est un example de décodage auto-spéculatif. \\
 \hline
\end{tabular}

\caption{An example of Simultaneous Translation IO Stream via re-translation}
\label{tab:example}
\end{table*}

\section{Method}
\label{sec:method}

\subsection{Streaming Re-translation}

We consider a streaming translation scenario where the source input arrives incrementally. At streaming step $T$, the model receives a partial source input $X^T$ and produces a target output $Y^T = (y_0, \dots, y_{n-1})$. Under autoregressive decoding:
\begin{equation}
Y^T = \prod_{i=0}^{n-1} F_{\text{decode}}\bigl(P(y_i \mid X^T, Y^T_{<i})\bigr),
\label{eq:decoder_equation_short}
\end{equation}
where $F_{\text{decode}}$ denotes a sampling or decoding function.

Re-translation applies this decoding procedure repeatedly as new source tokens arrive. While effective, this strategy recomputes the entire target prefix at each update, resulting in substantial redundant computation.

\subsection{Self-Speculative Decoding}

A key observation in streaming translation is that successive outputs typically share long common prefixes. Let $Y^{T-1}$ denote the translation produced at the previous streaming step, consisting of $m$ tokens. We treat $Y^{T-1}$ as a speculative draft for the current input $X^T$ and verify it in parallel rather than regenerating it autoregressively.

Specifically, draft tokens are verified in a single forward pass until the first mismatch at position $d$ ($d \le m$). Decoding then resumes autoregressively from that position:

\begin{equation}
    \begin{split}
    Y^T \approx
    \prod_{i=0}^{d-1} F_{\text{verify}}\bigl(P(y_i \mid X^T, Y^{T-1}_{<i})\bigr) \\
    \times \prod_{i=d}^{n-1} F_{\text{decode}}\bigl(P(y_i \mid X^T, Y^T_{<i})\bigr),
    \end{split}
    \label{eq:approximation}
\end{equation}
where $F_{\text{verify}}$ accepts draft tokens as long as they remain consistent with the current input.



\subsection{Biased Draft Verification}


Draft acceptance governs both speed and perceived flicker in streaming translation. To improve draft acceptance, we introduce a biased verification strategy that softly favors the draft token during verification. Given the model’s predicted distribution $P(\cdot)$, we construct a biased distribution:
\begin{equation}
    \begin{aligned}
    P'(y_i \mid X^T, Y^{T-1}_{<i}) 
    &= (1-\beta)\, P(y_i \mid X^T, Y^{T-1}_{<i}) \\
    &\quad + \beta\, \delta(y_i, Y^{T-1}_i),
    \end{aligned}
    \label{eq:biased_sampling}
\end{equation}
where $\delta(\cdot)$ assigns probability mass to the draft token and $\beta \in [0,1]$ controls the bias strength.
With a moderate bias value, this strategy increases the likelihood of retaining previously generated tokens when the model’s confidence is comparable while also limiting error accumulation as additional source context becomes available in future steps.

\subsection{Display-only Mask-k}

Previous work \citep{arivazhagan-etal-2020-translation,9054585} suggests masking the last $k$ tokens of intermediate outputs to reduce flickering. While effective, directly masking the output sequence removes reusable tokens from draft and limits potential of SSBD. We propose to apply mask-$k$ only at the display level, hiding unstable suffixes from the user while retaining them internally as draft tokens. This display-only strategy preserves decoding efficiency while significantly reducing perceptible flicker.



\begin{table*}[ht]
\begin{center}
{
\renewcommand{\arraystretch}{0.85}
\setlength{\tabcolsep}{4pt}

\begin{tabular}{l cccccc cccccc}
\toprule
\multirow{2}{*}{Method} &
\multicolumn{6}{c}{Tower+ 2B} &
\multicolumn{6}{c}{Qwen3 4B} \\
\cmidrule(lr){2-7} \cmidrule(lr){8-13}
 & {COMET} & {NE} & {A/D} & {A/O} & {TPS} & {Speedup} &
   {COMET} & {NE} & {A/D} & {A/O} & {TPS} & {Speedup} \\
\midrule
\multicolumn{13}{l}{\bfseries English$\rightarrow$German}\\[-0.6ex]
RT   & 0.877 & 1.17 & {--}  & {--}  & 60  & 1.00 & 0.843 & 1.26 & {--}  & {--}  & 48  & 1.00 \\
SSBD & 0.877 & 0.77 & 79.0  & 63.1  & 101 & 1.69 & 0.843 & 0.82 & 74.7  & 58.9  & 74  & 1.53 \\
\midrule
\multicolumn{13}{l}{\bfseries English$\rightarrow$Chinese}\\[-0.6ex]
RT   & 0.882 & 1.72 & {--}  & {--}  & 59  & 1.00 & 0.882 & 1.54 & {--}  & {--}  & 46  & 1.00 \\
SSBD & 0.880 & 1.01 & 71.7  & 57.0  & 87  & 1.48 & 0.882 & 1.20 & 65.4  & 51.8  & 65  & 1.43 \\
\midrule
\multicolumn{13}{l}{\bfseries English$\rightarrow$Japanese}\\[-0.6ex]
RT   & 0.912 & 2.06 & {--}  & {--}  & 60  & 1.00 & 0.889 & 1.96 & {--}  & {--}  & 48  & 1.00 \\
SSBD & 0.911 & 1.49 & 60.4  & 48.5  & 82  & 1.36 & 0.888 & 1.49 & 57.7  & 45.9  & 64  & 1.33 \\
\bottomrule
\end{tabular}
}
\end{center}
\caption{Overall performance on Flores in different translation directions comparing \textit{self-speculative biased decoding} (SSBD) with \textit{Re-translation} (RT). SSBD result is reported with bias \( \beta \) set to 0.2.}
\label{tab:main_results}
\end{table*}

A detailed algorithmic description of SSBD is provided in Appendix~\ref{sec:selfSpecBias}.

\section{Experiment}
\subsection{Setup}
\paragraph{Dataset} 
We evaluate SSBD on two benchmarks: the Flores text translation benchmark \citep{goyal-etal-2022-flores} and the ACL 60/60 speech translation evaluation set \citep{salesky-etal-2023-evaluating}. For Flores, we simulate streaming input by incrementally revealing source text using a fixed lag of $w=3$ words. For ACL 60/60, we generate streaming English transcripts using the Azure real-time ASR API and use intermediate transcriptions as incremental inputs for text-to-text simultaneous translation. Utterances split by long pauses are excluded to avoid manual post-editing.
We evaluate English-to-German, English-to-Chinese, and English-to-Japanese translation to cover varying linguistic distances.

\paragraph{Models} Experiments are conducted on two open-weight LLMs: Qwen3-4B \citep{yang2025qwen3} and Tower+ 2B \citep{rei2025towerplus}. All methods use \texttt{llama.cpp} on Apple Silicon (Metal).

\paragraph{Metrics}
Translation quality is measured using COMET \citep{rei-etal-2022-comet}. Output stability is evaluated using Normalized Erasure (NE) \citep{arivazhagan-etal-2020-translation,9054585}, computed over tokenized outputs. Decoding efficiency is reported in output tokens per second (TPS) and relative speedup over autoregressive re-translation. We additionally report draft acceptance metrics, including accepted draft tokens over draft length (A/D) and over total output tokens (A/O).

\subsection{Results}
\paragraph{Overall performance}
Table~\ref{tab:main_results} presents Flores dataset results, where SSBD consistently outperforms re-translation from scratch, achieving 1.3--1.7$\times$ speedups with comparable COMET scores across models and languages. Additional results on the ACL 60/60 dataset are reported in Appendix~\ref{sec:acl_6060} and exhibit similar trends.

\paragraph{Effect of Language Pair}
Speedup varies across language pairs, reflecting differences in prefix stability under incremental inputs. English-to-Japanese exhibits higher NE and lower draft acceptance due to substantial word-order divergence, while English-to-German and English-to-Chinese exhibit higher prefix stability, resulting in greater speedup.

\paragraph{Effect of Biased Verification}
Table~\ref{tab:bias_effect} illustrates the performance trade-offs associated with the bias parameter $\beta$. We observe that increasing $\beta$ improves draft acceptance rates (A/D) and reduces the NE, accelerating decoding. However, an excessively high bias leads to a decline in COMET scores by constraining the model's ability to revise earlier tokens when additional source context arrives. In the limit, the system converges toward \textit{Continuous Decoding (CD)}, maximizing throughput but substantially degrading quality due to unrectified early-stage errors. Consequently, we fix $\beta = 0.2$ for all primary experiments to maintain an optimal balance between latency and translation accuracy.

\begin{table}[htbp]
\centering

\renewcommand{\arraystretch}{0.85}
\begin{tabular}{rccccc}
    \hline
    $\beta$ & COMET & NE & A/D & A/O & Speedup \\
    \hline
    \multicolumn{6}{l}{\small \bf Flores English$\rightarrow$Chinese}\\
    RT  & 0.882 & 1.72 & -      & -     & 1.00 \\
    \hdashline
    0.0 & 0.882 & 1.53 & 53.8   & 42.7  & 1.28 \\
    0.1 & 0.881 & 1.19 & 66.2   & 52.6  & 1.40 \\
    0.2 & 0.880 & 1.01 & 71.7   & 57.0  & 1.48 \\
    0.3 & 0.870 & 0.83 & 77.3   & 61.4  & 1.58 \\
    0.4 & 0.842 & 0.57 & 85.0   & 67.4  & 1.70 \\
    0.5 & 0.742 & 0    & 100.0  & 77.5  & 2.10 \\
    1.0 & 0.742 & 0    & 100.0  & 77.5  & 2.10 \\
    \hdashline
    CD  & 0.742 & 0    & 100.0  & 77.5  & 2.60 \\
    \hline
\end{tabular}

\caption{The effect of bias \( \beta \) in Biased Draft Verification. An excessively high bias converges toward the \textit{Continuous Decoding} (CD).}
\label{tab:bias_effect}
\end{table}

\paragraph{Display-only Mask-$k$}
We evaluate the effect of applying mask-$k$ in Table~\ref{tab:mask_k}. While masking intermediate output suffix tokens directly reduces flickering measured by NE, it also reduces the share of accepted tokens in the output sequence (A/O) due to the removal of those potentially acceptable draft tokens. This limit can be lifted by applying \textit{display-only mask-$k$}, preserving acceptable draft tokens for speedup while reducing flickers from user perception.

\begin{table}[htbp]
\centering

\renewcommand{\arraystretch}{0.85}
\begin{tabular}{rccccc}
    \hline
    k & COMET & NE & A/D & A/O & Speedup \\
    \hline
    \multicolumn{6}{l}{\small \bf Flores English$\rightarrow$Chinese}\\
    - & 0.880 & 1.01 & 71.7   & 57.0  & 1.48 \\
    \hdashline
    \multicolumn{6}{l}{\small \bf +mask-k}\\
    3 & 0.880 & 0.52 & 80.6   & 52.3  & 1.42 \\
    5 & 0.881 & 0.34 & 84.5   & 47.2  & 1.39 \\
    \hdashline
    \multicolumn{6}{l}{\small \bf +display-only mask-k}\\
    3 & 0.880 & 0.53 & 71.7   & 57.0  & 1.48 \\
    5 & 0.880 & 0.35 & 71.7   & 57.0  & 1.48 \\
    \hline
\end{tabular}

\caption{Output mask-$k$ removes potentially acceptable tokens from the draft, limiting the potential of SSBD.}
\label{tab:mask_k}
\end{table}

\section{Conclusion}

We propose \textit{Self-Speculative Biased Decoding} (SSBD), a simple yet effective inference paradigm designed to accelerate LLMs in streaming scenarios. By reusing the previous output as a draft and applying probability biasing during draft verification, SSBD reduces redundant computation and increases draft acceptance rates, leading to faster inference and significantly reduced flickering. We further introduce a \textit{display-only mask-$k$} heuristic that preserves reusable draft tokens while minimizing perceptible flicker. Experiments on simultaneous translation show that SSBD achieves substantial speedups without sacrificing translation quality or output stability.

\section*{Limitations}
Although SSBD is deployable out of the box without architectural changes or training, its effectiveness depends on the monotonicity of model outputs under partial inputs. Models not explicitly trained for prefix monotonicity may exhibit lower draft acceptance rates. While prefix training could improve monotonicity and further enhance SSBD, this remains beyond the scope of this work. Moreover, SSBD is evaluated primarily on simultaneous translation; its effectiveness in other streaming generation tasks (e.g., summarization or ASR post-editing) has not yet been empirically validated.

\bibliography{anthology-cited,custom}

\appendix

\section{Appendix}
\label{sec:appendix}
\subsection{Experiment details}
Model artifacts used throughout the paper are downloaded from their official Hugging Face repos under the names \textit{Unbabel/Tower-Plus-2B} and \textit{Qwen/Qwen3-4B}. Both are exported to GGUF format and quantized into \textit{Q4\_K\_M} using \texttt{llama.cpp}.

Inference is conducted on an Apple M2 Pro
processor with the Metal backend using greedy search during decoding. For each setting, we report the mean TPS over three runs.

We report the COMET score using the default model \textit{Unbabel/wmt22-comet-da}, and NE is computed on the output tokenized by the Flores-101 tokenizer from SacreBLEU. Output tokens per second (TPS) are measured as the number of output tokens emitted per update divided by wall-clock generation time; for SSBD the timing includes the draft-verification forward pass. We present the prompt templates used throughout the experiments in Table~\ref{tab:prompts}.

Code to reproduce our experiments will be released upon publication, including scripts to generate streams, evaluation, and decoding implementation.

\subsection{Algorithmic Description}\label{sec:selfSpecBias}
This section provides a detailed algorithmic description of Self-Speculative Biased Decoding (SSBD) for completeness in Algorithm~\ref{alg:selfSpecBias}.
The pseudocode follows the method described in Section~\ref{sec:method} and clarifies the draft reuse and biased verification process at each streaming step.

\subsection{Performance on ACL 60/60}
\label{sec:acl_6060}
The overall performance of SSBD compared with re-translation tested on ACL 60/60 speech translation evaluation set is presented in the Table~\ref{tab:acl_6060}. A similar trend to what was observed in Flores can be seen here as well.

\subsection{Case studies}
Table~\ref{tab:case_study} presents a representative example illustrating the extent to which SSBD reduces redundant computation by reusing previously generated output tokens instead of re-decoding them from scratch. The example highlights that, with a modest bias, the model is able to retain drafted tokens when their predicted confidence remains comparable, while still correcting earlier suboptimal predictions as additional source context becomes available. In contrast, an excessively large bias may cause the model to overcommit to prior predictions, reducing its ability to revise previously generated tokens when corrections are warranted.

\begin{algorithm}
\caption{Self-Speculative Biased Decoding}
\label{alg:selfSpecBias}
{\raggedright
\textbf{Input}: Current input prefix $X^T = (x_0, ..., x_{n-1})$, previous output $Y^{T-1} = (y_0, ..., y_{m-1})$, model $p_{\theta}$ with KV cache \\
\textbf{Parameter}: bias $\beta$ \\
\textbf{Output}: output $Y^T$ for input $X^T$ \par}
\begin{algorithmic}[1] 
\STATE Let draft sequence $D \gets Y^{T-1}$
\STATE Let $prompt \gets \text{create\_input}(X^T,\texttt{<BOS>})$, run forward pass with cached states for it
\STATE Run one forward pass over the draft tokens D, obtaining output probabilities $p_i = p_{\theta}(\cdot| X_t, y_0, ..., y_{i-1})$ for $i=0, ..., m-1$
\STATE Let diverged position $d \gets 0$.
\FOR{$i = 0$ \TO $m-1$}
    \STATE $p'(i) \gets p(i) \cdot (1-\beta)+\beta \cdot \delta(\cdot=y_i)$
    \IF {$y_i \ne \operatorname{argmax}(p'(i))$}
        \STATE \textbf{break}
    \ENDIF
    \STATE {d++}
\ENDFOR
\STATE Accepted tokens $D_a \gets y_0, ..., y_{d-1}$
\STATE Init decoding position $j \gets d$, clear all cache states from position $j$

\WHILE{Stopping criteria not met}
\STATE {Run step-wise model forward to sample $y_j$;} \\
\STATE {j++}
\ENDWHILE

\STATE \textbf{return} Concat of accepted tokens and newly decoded suffix as $Y^T$
\end{algorithmic}

\end{algorithm}

\begin{table*}[ht]
\centering
\centering
\footnotesize
\begin{tabular}{l}
\toprule
\textbf{Prompt Template} \\
\midrule
\textbf{Unbabel/Tower-Plus-2B} \\

\begin{minipage}{0.95\linewidth}
\begin{Verbatim}[fontsize=\footnotesize]
<start_of_turn>user
Translate the {{source language}} source text to {{target language}}. Return only the translation,
without any additional explanations or commentary.
{{source language}}: {{User Input}}
{{target language}}: <end_of_turn>
<start_of_turn>model
\end{Verbatim}
\end{minipage}
\\
\midrule
\textbf{Qwen/Qwen3-4B} \\

\begin{minipage}{0.95\linewidth}
\begin{Verbatim}[fontsize=\footnotesize]
<|im_start|>system
Translate the {{source language}} source text to {{target language}}. Return only the translation,
without any additional explanations or commentary.<|im_end|>
<|im_start|>user
{{source language}}: {{User Input}}<|im_end|>
<|im_start|>assistant
<think>

</think>

{{target language}}:
\end{Verbatim}
\end{minipage}
\\
\bottomrule
\end{tabular}

\caption{Prompts used throughout the experiments. Since the \{\{User Input\}\} is updated at every streaming step, prefix caching across steps is limited to the shared prompt and the shared source-prefix tokens; tokens after the assistant start position generally cannot be reused across steps. Output TPS is counted from the first token of model output, after the prompt templates shown above.}
\label{tab:prompts}
\end{table*}

\begin{table*}[ht]
\begin{center}
{
\setlength{\tabcolsep}{4pt}

\begin{tabular}{l cccccc cccccc}
\toprule
\multirow{2}{*}{Method} &
\multicolumn{6}{c}{Tower+ 2B} &
\multicolumn{6}{c}{Qwen3 4B} \\
\cmidrule(lr){2-7} \cmidrule(lr){8-13}
 & {COMET} & {NE} & {A/D} & {A/O} & {TPS} & {Speedup} &
   {COMET} & {NE} & {A/D} & {A/O} & {TPS} & {Speedup} \\
\midrule
\multicolumn{13}{l}{\bfseries English$\rightarrow$German}\\[-0.8ex]
RT   & 0.796 & 2.53 & {--}  & {--}  & 60  & 1.00 & 0.766 & 2.22 & {--}  & {--}  & 48  & 1.00 \\
SSBD & 0.796 & 1.82 & 76.3  & 55.3  & 90  & 1.50 & 0.766 & 1.79 & 73.1  & 52.1  & 70  & 1.46 \\
\midrule
\multicolumn{13}{l}{\bfseries English$\rightarrow$Chinese}\\[-0.8ex]
RT   & 0.824 & 2.34 & {--}  & {--}  & 59  & 1.00 & 0.833 & 2.15 & {--}  & {--}  & 46  & 1.00 \\
SSBD & 0.824 & 1.71 & 74.5  & 53.8  & 84  & 1.42 & 0.833 & 1.84 & 67.5  & 48.5  & 59  & 1.28 \\
\midrule
\multicolumn{13}{l}{\bfseries English$\rightarrow$Japanese}\\[-0.8ex]
RT   & 0.854 & 3.05 & {--}  & {--}  & 59  & 1.00 & 0.841 & 2.91 & {--}  & {--}  & 44  & 1.00 \\
SSBD & 0.854 & 2.39 & 60.9  & 42.5  & 78  & 1.31 & 0.841 & 2.46 & 56.3  & 40.8  & 61  & 1.39 \\
\bottomrule
\end{tabular}
}
\end{center}
\caption{Overall performance on ACL 60/60 in different translation directions comparing \textit{self-speculative biased decoding} (SSBD) with \textit{Re-translation} (RT). SSBD result is reported with bias \( \beta \) set to 0.2. We can identify a similar trend as observed previously with the Flores testset.} 
\label{tab:acl_6060}
\end{table*}




\begin{table*}[ht]
\centering
\centering
\footnotesize
\begin{CJK*}{UTF8}{gbsn}
\begin{tabular}{p{0.95\linewidth}}
\toprule
\textbf{Incremental Translation Outputs} \\ \midrule
\textbf{English Input} \\
\textbf{1/3 gap in} \\
\underline{1/3 gap in}\textbf{ research is that} \\
\underline{1/3 gap in research is that}\textbf{ new pre trained language} \\
\underline{1/3 gap in research is that new pre trained language }\textbf{models} \\
\underline{1/3 gap in research is that new pre trained language models }\textbf{are most often} \\
\underline{1/3 gap in research is that new pre trained language models are most often }\textbf{evaluated on} \\
\underline{1/3 gap in research is that new pre trained language models are most often evaluated on }\textbf{high resource} \\
\underline{1/3 gap in research is that new pre trained language models are most often evaluated on high resource }\textbf{languages.} \\

\midrule
\textbf{Re-translation Baseline (RT)} \\
\textbf{差距为1/3} \\
\textbf{研究 中的 三分 之一 差距 是} \\
\texttt{研究 中}\textbf{存在 三分 之一 的 空白 是 新 预训练 语言} \\
\texttt{研究 中 存在}\textbf{的 三分 之一 的 差距 是 新型 预训练 语言模型} \\
\texttt{研究 中 存在 的 三分 之一 的 差距}\textbf{ 在于 , 新训练好的 语言模型 通常} \\
\texttt{研究 中 存在 的 三分 之一}\textbf{ 差距 在于 , 新训练好的 语言模型 通常是以以下方式进行评估的} \\
\texttt{研究 中 存在 的}\textbf{ 1/3 差距 在于 , 新训练好的 语言模型 通常 都是 在 高 资源环境下 进行 评估的。} \\
\texttt{研究 中 存在 的 1/3 差距 在于, 新}\textbf{型预训练语言模型通常都是在高资源语言上进行评估。} \\

\midrule
\textbf{SSBD with modest bias} ($\beta = 0.2$) \\
\textbf{差距为1/3} \\
\textbf{研究中的三分之一差距是} \\
\underline{研究中的三分之一差距是}\textbf{新预先训练的语言} \\
\underline{研究中的三分之一差距是新预先训练的语言}\textbf{模型} \\
\underline{研究中的三分之一差距是新预先训练的语言模型}\textbf{通常} \\
\underline{研究中的三分之一差距}\textbf{在于,新训练好的语言模型通常是通过以下方式进行评估的} \\
\underline{研究中的三分之一差距在于,新训练好的语言模型通常}\textbf{是通过高资源来评估的。} \\
\underline{研究中的三分之一差距在于,新训练好的语言模型通常是}\textbf{以高资源语言进行评估的。} \\

\midrule
\textbf{SSBD with overly large bias} ($\beta \ge 0.5$) or \textbf{Continuous Decoding} \\
\textbf{差距为1/3} \\
\underline{差距为1/3}\textbf{的研究是因为} \\
\underline{差距为1/3的研究是因为}\textbf{新预测语言} \\
\underline{差距为1/3的研究是因为新预测语言}\textbf{模型} \\
\underline{差距为1/3的研究是因为新预测语言模型}\textbf{通常} \\
\underline{差距为1/3的研究是因为新预测语言模型通常}\textbf{都需要经过评估} \\
\underline{差距为1/3的研究是因为新预测语言模型通常都需要经过评估}\textbf{高资源} \\
\underline{差距为1/3的研究是因为新预测语言模型通常都需要经过评估高资源}\textbf{语言。} \\

\bottomrule
\end{tabular}
\end{CJK*}
\caption{Incremental translation comparison. (1) The underlined portion of the English input represents the prefix cached text, bold text represents newly added input. (2) Re-translation baseline re-decode all output tokens and frequently rephrases or alternates the choice of words, even not necessary; (3) SSBD is most likely to reuse previous decoded tokens as underlined, and continue to generate new token shown in bold as needed; (4) SSBD with modest bias can still allow the model to rewrite its previous prediction when necessary; (5) SSBD with overly large bias promotes too much confidence in the previous predicted token, preventing necessary rewriting. In this regime, the behavior converges toward \textit{Continuous Decoding}, maximizing stability at the cost of error recovery and overall translation quality.}

\label{tab:case_study}
\end{table*}

\end{document}